# Learning Where to See: A Novel Attention Model for Automated Immunohistochemical Scoring

Talha Qaiser, Nasir M. Rajpoot, *Senior Member*, *IEEE*

*Abstract*—Estimating over-amplification of human epidermal growth factor receptor 2 (HER2) on invasive breast cancer (BC) is regarded as a significant predictive and prognostic marker. We propose a novel deep reinforcement learning (DRL) based model that treats immunohistochemical (IHC) scoring of HER2 as a sequential learning task. For a given image tile sampled from multi-resolution giga-pixel whole slide image (WSI), the model learns to sequentially identify some of the diagnostically relevant regions of interest (ROIs) by following a parameterized policy. The selected ROIs are processed by recurrent and residual convolution networks to learn the discriminative features for different HER2 scores and predict the next location, without requiring to process all the sub-image patches of a given tile for predicting the HER2 score, mimicking the histopathologist who would not usually analyze every part of the slide at the highest magnification. The proposed model incorporates a task-specific regularization term and inhibition of return mechanism to prevent the model from revisiting the previously attended locations. We evaluated our model on two IHC datasets: a publicly available dataset from the HER2 scoring challenge contest and another dataset consisting of WSIs of gastroenteropancreatic neuroendocrine tumor sections stained with Glo1 marker. We demonstrate that the proposed model outperforms other methods based on state-of-the-art deep convolutional networks. To the best of our knowledge, this is the first study using DRL for IHC scoring and could potentially lead to wider use of DRL in the domain of computational pathology reducing the computational burden of the analysis of large multi-gigapixel histology images.

*Index Terms* — Deep Reinforcement Learning, Computational Pathology, Immunohistochemical Scoring, Breast Cancer.

## I. INTRODUCTION

Human epidermal growth factor receptor 2 (HER2) is a protein that influences the growth of malignant epithelial cells. The over-amplification of HER2 gene is observed in breast cancer (BC) cases having invasive tumor regions. Cancer cells in HER2+ (positive) cases of BC encounter neoplastic transformations which lead to an uncontrolled growth of tumor cells as compared to HER2− (negative) cases. Approximately all the invasive breast carcinomas cases are recommended for HER2 testing [1] and nearly 20-30% of cases have overexpression of HER2 protein [2] which is associated with poor prognosis, lower survival, and high recurrence [3]. Recent studies have reported the HER2 status as a predictive factor for anti-HER2 and hormonal therapies and also a prognostic factor to associate invasive tumors with mortality and duration of recurrence free survival [4]. Therefore, precise quantification of HER2 overexpression is crucial for ensuring that HER2+ patients receive appropriate anti-HER2 treatment.

The most common method for assessment of HER2 biomarker expression is immunohistochemical (IHC) staining. In routine clinical practice, an expert pathologist visually examines IHC stained BC histology slides under the microscope to quantify the expression of HER2 and reports a score between 0 and 3+. Example regions of interest (ROIs) from slides with different HER2 scores (0 to 3+) are shown in Fig. 1. Samples scoring 0 or 1+ contain no or weak membrane staining in less than 10% of the tumor cells and are regarded as HER2−. A score of 3+ is assigned to cases where strong membrane staining is observed in more than 10% of the tumor cells and regarded as HER2+. Borderline cases, in which a non-uniform staining is observed, are classified as equivocal with a 2+ score. Such cases are recommended for fluorescence in-situ hybridization (FISH) test to measure the status of HER2/neu gene amplification. A limitation of the current practice for HER2 scoring is that the visual examination of glass slides is a time consuming and laborious task, as well as subjective by nature and more likely to be affected by inter- and intra-observer variability. It has also been reported that up to 20% of the HER2 results may contain inaccuracies [5].

Recently, there has been an upward trend in the adoption of digital pathology and consequently a surge of research on algorithms for analysis of pathology images. Despite the recent progress that highlights the significance of image analysis in computational pathology [6][7], there are several challenges that hinder the adoption of algorithms in routine clinical practice. Computational pathology algorithms usually require detailed annotated datasets to predict the slide label. For the task at hand, the ground-truth (GT) label for IHC score (HER2 score in our case) is generally provided at the whole slide image (WSI) level and there are no detailed annotations provided about which ROIs from the tissue slides are consulted for the final HER2 score. Amongst existing automated approaches, the simplest approach is to manually or randomly extract patches from desired ROIs of a WSI and train a supervised model to predict the required HER2 score.

T. Qaiser is with Department of Computer Science, University of Warwick, Coventry, UK (e-mail: t.qaiser@warwick.ac.uk).
N. M. Rajpoot is with Department of Computer Science, University of Warwick, Coventry, UK, Department of Pathology, University Hospitals Coventry and Warwickshire, Coventry, UK and The Alan Turing Institute, London, UK (e-mail: n.m.rajpoot@warwick.ac.uk).

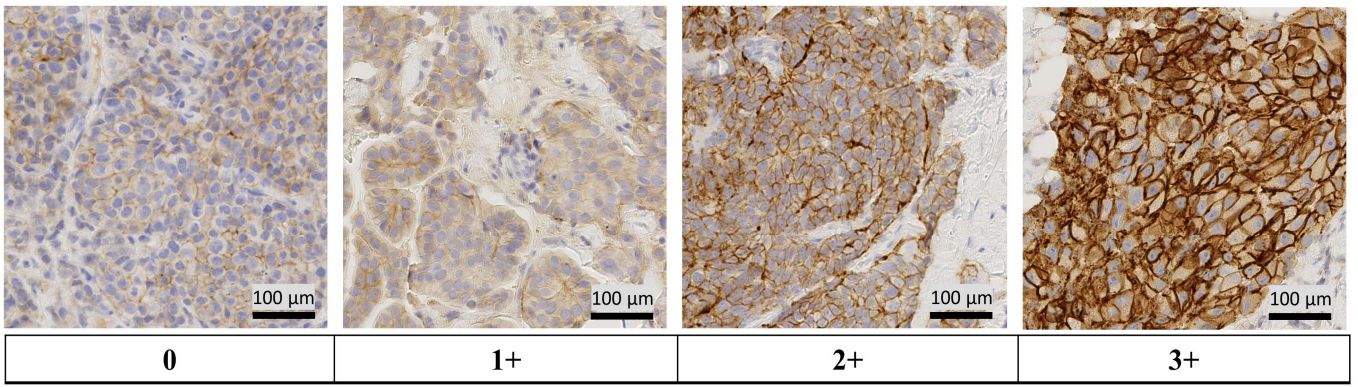

Fig. 1: *Left to right*: examples of regions of interest from whole-slide images of the training dataset.

Such approaches offer an inevitable bias to the model predictions and disjoint selection of small patches may also suffer from loss of visual context. Another potential shortcoming is computational inefficiency, as these models need to process all the regions of a given image, where some of the tissue regions may not be diagnostically relevant for the prediction of the correct HER2 score.

With regards to the aforementioned challenges, we ask the question: can we train a model that ignores the irrelevant information and learn *where to see*? To answer this question, we propose a novel deep learning approach for automated scoring of IHC stained HER2 slides of invasive breast carcinoma, based on the concept of policy gradients. Given a large tile from a WSI, the proposed model identifies some of the diagnostically relevant locations from a low resolution (2.5 ×) coarse representation of a given image by learning a parameterized policy over the interaction sequences of ROIs locations. The model sequentially samples the multi-resolution ROIs 40 × and 20 ×, from the relevant locations to learn the discriminative features for different HER2 scores (0 to 3+). The core components of the proposed model are a residual convolutional neural network (ResNet) and a recurrent neural network (RNN). The role of ResNet in this model is to learn discriminative features whereas the RNN sequentially analyzes the provided features to predict the outcome and the next location. Since the GT information was only provided for WSI level with no prior knowledge of ROI locations, we train the model with policy gradients. Our model is designed to explore spatially distinct locations and learn features from visually discriminative regions. In cognitive psychology, this phenomenon is known as inhibition of return (IoR) [8] that prevents the previously attended regions to be attended again. Our model incorporates the concept of IoR in order to encourage the model to attend non-overlapping diagnostically relevant regions. Another important issue is that an erroneous scoring of 3+ as 0/1+ or vice versa may have far-reaching effects for a patient. In order to avoid such large errors, we propose a task-specific regularization term that penalizes such predictions. This study was conducted on a publicly available dataset from the HER2 scoring contest [9] containing 172 WSIs from 86 cases. Extensive experiments on the contest dataset show the efficacy of our proposed model, for guiding deep learning models to ignore irrelevant regions and scaling up to large images. The proposed method outperforms all the 18 algorithms that participated in the HER2 contest, most methods using state-of-the-art CNNs.

## II. RELATED WORK

Automated IHC scoring has been approached with a variety of handcrafted features and deep learning based methods. The most common approach for automated IHC scoring involves a pre-processing step to identify the potential tissue regions for training the underlying model. Then, a handful of small patches are sampled from selected tissue regions, either randomly or by using sliding window approach. The identification of potential tissue regions is generally accomplished by manual selection [10][11], semi-automated [12] or thresholding based automated methods [13]. The pre-processing step is generally followed by training a patch-based supervised model, to learn the discriminative features and predict the outcome of each input patch. A range of hand-crafted [13][14], approaches have been proposed to improve the IHC scoring of hormone receptors in breast cancer. For HER2 scoring, Rodner *et al.* [15] recently proposed an algorithm that computes a set of bilinear filters using convolutional layers. For classification of HER2 scores on patch level, they use bilinear features to train a multi-class logistic regression model. A deep neural network has been presented by Saha *et al.* [16] for HER2 quantification by segmenting nuclei and cell membranes. Mukundan *et al.* [17] introduced a set of characteristic curves by varying the intensity of saturation channel with a handpicked threshold for classification of HER2 score. The final step of HER2 scoring in general involves aggregation of patch level scores to the WSI level score which is typically done by finding the most dominant class within a WSI or by training a shallow classifier on features selected from the output probability map of a WSI. Supervised patch-based approaches have established well for problems where tissue level GT is readily available. However, in IHC scoring where tissue level GT is generally not available, it is imperative to explore how deep learning models can be trained to ignore unnecessary information from the given image and focus only on regions that eventually helps in predicting the correct outcome.

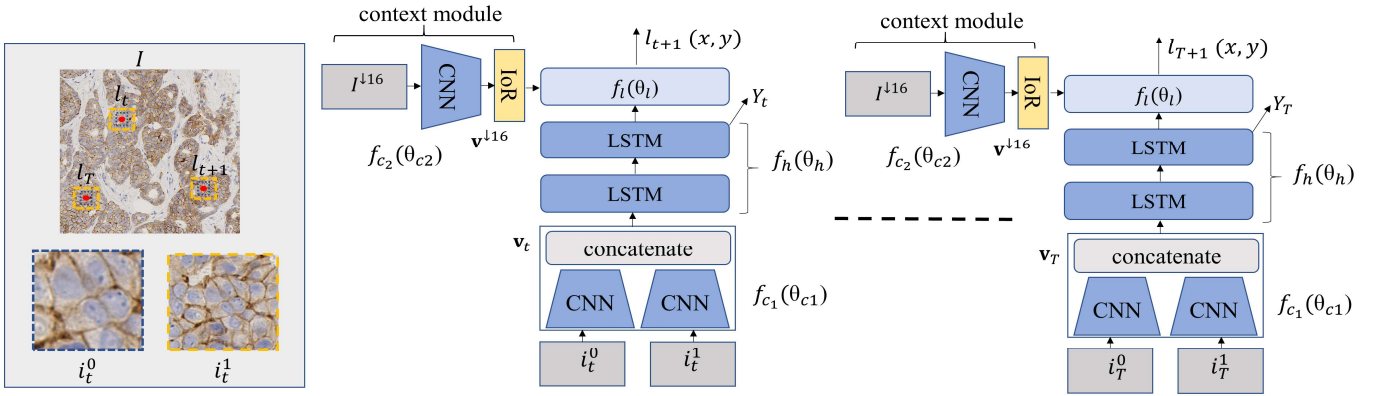

Fig. 2: The schematic diagram of our deep recurrent model. The regions of interest sampled around $l_t$ at 40× and 20× are shown with blue and yellow dashed bounding boxes, respectively. The dashed line in the recurrent model shows the **T** sequential iterations. IoR act as a penalization term in the loss function.

Recent studies have shown that deep reinforcement learning (DRL) has been employed in widespread applications. For object detection, Caicedo *et al.* [18] proposed a deep Q-Network (DQN) for multi-class object localization. The model localizes target objects by following a search strategy, which starts with analyzing the input image and then agent guides the model to narrow down the field of view for precise object localization. The reward function was calculated by computing the intersection-over-union between the GT and the predicted bounding box for the object. This work was further extended for medical images including automated anatomical landmark [19] and breast lesion detection [20] for DCE-MRI images, whereby an agent localizes the potential ROI containing the lesion by iteratively adjusting the bounding box. The reward function was computed by using the Dice coefficient between the GT and the predicted box. There also exist some works that incorporate DRL for genomics data to enhance the annotation of biological sequences in genome sequencing [21] and construction of protein interaction network for prostate cancer [22]. Another interesting extension is the incorporation of attention models with policy gradients that enable the underline model to learn a parametrized policy based on spatial dependencies. This combination has been explored for a number of applications including object detection [23], action recognition [24] and image captioning [25]. However, having precisely annotated GT for computing the reward function limits the use of deep Q-learning for IHC scoring.

In this work, we treat IHC scoring as a sequential learning task to learn discriminative features and select informative regions within a large image tile of a WSI. To the best of our knowledge, this is the first study that uses DRL for IHC scoring of histology WSIs. In terms of policy learning methodology, our work has some similarities to the methods proposed by Mnih *et al.* [26] and Ranzato *et al.* [27]. Our proposed model contains a context module that incorporates the coarse representation of input image, before predicting attentive locations. The end-to-end inhibition of return (IoR) mechanism encourages the model to explore spatially distinct attentive locations. Moreover, the scope of existing attention methods is limited to relatively small natural images whereas tumors in IHC stained WSIs are heterogeneous in terms of their morphological appearance, color variability, shape, and temporal locations.

### III. LEARNING WHERE TO SEE

Given an image $I$, the task is to predict the HER2 score ranging from 0 to 3+ by selecting a set of diagnostically relevant regions as well as learning discriminative features from those regions. The schematic diagram of the proposed model is shown in Fig. 2. At each time step $t$, the model receives two ROIs $i_t = (i_t^0, i_t^1)$, where $i_t^0, i_t^1 \subset I$ are regions of width 128 and height 128 sampled at the region center $l_t$ at different magnification levels 40× and 20×, respectively. The convolutional network $f_{c1}$ with learnable parameters $\theta_{c1}$ analyzes $i_t$ and transforms it into a fixed length feature vector $\mathbf{v}_t \in \Re^m$. The recurrent model $f_h$ with learnable parameters $\theta_h$ sequentially processes the aggregated ROI features to update its internal state. Besides, the context model processes the down-sampled version $I^{\downarrow 16}$ (down-sampled by a factor of 16 in both directions) of the input image and perform the IoR operation, as described in Section III.D. The next location $l_{t+1}$ is predicted by analyzing the hidden state $(f_h(\theta_h))$ from the RNN that reflects *where we currently are,* and the output $\mathbf{v}^{\downarrow 16}$ of CNN $f_{c2}$, with learnable parameters $\theta_{c2}$, that represents *the context*. The whole process is repeated for $T$ iterations and at the end of the sequence $(i_1, i_2, \ldots, i_T)$, the model predicts the final output score $Y_T$.

This iterative process wrapped around an RNN model forms a classical *environment-agent interface* that can be formalized by the partially observable Markov decision process (POMDP). In the current setup, the RNN and CNNs collectively act as a decision maker, which is formally known as an *agent* in the reinforcement learning (RL) literature. The agent sequentially interacts with the *environment*, which in our case is the image $I$. For each time step $t$, the agent receives a state from the environment. It then processes the given state and responds with appropriate actions, which in our cases is the next location ($l_{t+1}$), directing the model *where to see* and eventually deciding on the HER2 score. Overall this process of predicting the next location is partly stochastic and non-differentiable, requiring the use of policy gradients. The ultimate task for an agent is to map the given states into actions by learning a parametrized policy ($\pi$) through trial and error. At the end of this sequential process, the agent receives a scalar reward $R$ based on its actions, which in our case is linked to correct prediction of the HER2 score as described in Section III.C.

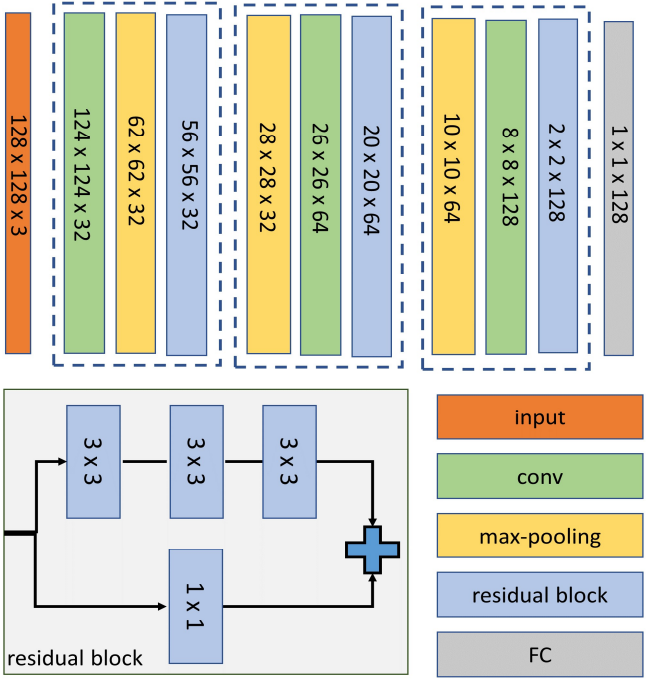

Fig. 3: A schematic illustration of residual convolution neural network.

At a high level, our model mimics the histopathologist practice treating a given image as an *environment* and the histopathologist as an *agent* who acts as a decision maker. Given the environment, the agent glances through different tissue components at a high level (low magnification) and then selects certain visual fields (ROIs) at low level (high magnification) to observe and store the relevant morphological features into the *memory*. The agent repeats this process for a certain number of iterations on different states to build an internal representation of the overall environment before coming up with the final decision, i.e. assigning the final score. For the sake of simplicity, in the remainder of this paper, we refer to the combination of RNN and CNNs as an agent and the selected ROIs as the state.

*A. Sequential Modelling*

The recurrent component (RNN) acts as the backbone of the proposed model. At each time step $t$, the recurrent model updates the parameters of hidden states and predicts the HER2 score for a CNN based feature representation of $i_t = (i_t^0, i_t^1)$. The input to the RNN is $\mathbf{v}_t$, the CNN based feature vector for multi-resolution ROIs $i_t$ centered at location $l_t$. The RNN model also updates internal states (*memory*) that capture information from previous time steps. The output $f_h(\theta_h)$ of the RNN model is defined as below,

$$f_{h^*}(\theta_{h^*}) = f(h_{t-1}, f_{c1}(\theta_{c1}); \theta_{h^*}) \quad (1)$$

$$f_h(\theta_h) = f(h_{t-1}, f_{h^*}(\theta_{h^*}); \theta_h) \quad (2)$$

We choose long short-term memory (LSTM) [28] as a preferred choice for RNN to learn spatial dependencies between the ROIs. LSTM has proven to be more robust and less likely to suffer from the problem of vanishing gradients. An important task of the model is to predict the next location $l_{t+1}$ by using $\mathbf{v}^{\downarrow 16}$ provided by CNN and hidden representation $f_h(\theta_h)$ of ROI images processed by CNN and LSTM. We computed the Hadamard product of $f_h(\theta_h)$ and $\mathbf{v}^{\downarrow 16}$ to obtain a combined feature vector. Finally, the location module $f_l(\theta_l)$, linearly transforms the combined representation to predict the normalized coordinates of the next location $l_{t+1}(x, y)$. During the training process, the model eventually learns to encode the information from the past sequences and decide *where to see*.

*B. Residual Convolutional Network*

In the proposed model, the convolutional network serves as a non-linear function that maps a given RGB image into a fixed length vector representation. More specifically, we are using a variant of residual CNN [29] that contain residual connections to reroute the input information into deeper convolutional layers. Residual blocks ensure the end-to-end training of deep models by preventing the gradient to vanish within lower layers of a CNN. It also enables the underlying model to reuse the low level features along with deeper convolutional (high level) features. For a given input $p_k$, the residual block function is defined as below

$$p_{k+1} = \sigma(F_k(p_k, w_k) + q(p_k)) \quad (3)$$

where $F_k(p_k, w_k)$ is representing a sequence of convolutional operations, $w_k$ denote the trainable weights (biases are omitted), $k$ represents the $k^{th}$ residual block of the CNN, $p_{k+1}$ is the output of residual block and $q(p_k) = p_k$ is an identity function. The function $\sigma(.)$ denotes a non-linear activation function, which in our case is a ReLU. The $F_k(p_k, w_k) + p_k$ operation is representing element-wise addition of two activation maps. A schematic illustration of the residual CNN is shown in Fig. 3.

We use two residual CNNs, $f_{c1}$ and $f_{c2}$ where $f_{c1}$ learns discriminative features by producing a fixed length non-linear vector representation $\mathbf{v}_t$ for further processing by the recurrent network and $f_{c2}$ incorporates the context information. The input to $f_{c1}(\theta_{c1})$ is the corresponding ROIs (at $40\times$ and $20\times$) sampled at $l_t(x, y)$ from the input image $I$. The network separately processes the selected ROIs and concatenates the feature representative of $i_t^0$, $i_t^1$. The concatenation operation is previously used in [30] for a hierarchical CNN to combine the feature representation of corresponding up/down convolution layers. The CNN features of extracted ROIs mainly assist our model to learn the discriminative patterns of different HER2 scores. The task for $f_{c2}(\theta_{c2})$ is to embed the contextual awareness within the proposed model.

*C. Model Training*

The proposed model (Fig. 2) is trained end-to-end by maximizing the performance over the parameters $\theta = (\theta_c, \theta_h, \theta_l)$ of the residual CNNs, recurrent network and the location module. By interacting with the environment, the model forms a special case of the POMDP framework with an episodic sequence of states, actions, and rewards. The task for the model is to learn a parameterized DRL policy ($\pi$) that maps a given state into action(s) by maximizing the sum of expected reward while following the parameterized policy $\pi$. The parameterized policy $\pi$ for calculating the probability of a certain action $a$ from the action space $A$ at iteration $t$ for a given state $s$ with parameters $\theta$ can be defined as follows,

$$\pi(a|s) = P(A_t|S_t; \theta) \quad (4)$$

where $S_t$ denotes the set of possible states at time $t$. Similarly, for the problem at hand, the model needs to learn a parameterized policy $\pi((l_{t+1}, Y_t) \mid (i_t, I^{\downarrow 16}); \theta)$ to predict the HER2 score ($Y_t$) and coordinates of the next location $l_{t+1}(x,y)$, given the selected ROIs ($i_t$) from $l_t(x,y)$ and down-sampled input image $I^{\downarrow 16}$. The model receives a scalar reward $r_t$ after interacting with each selected ROIs ($i_{1,2,...T}$) of the given image,

$$r_t = \begin{cases} 1 & g = \underset{s}{\mathrm{argmax}}\, Y'_T \\ 0 & \text{otherwise} \end{cases} \quad (5)$$

as provided in (5), where $g$ denotes the GT score, $Y'_T$ represents the output of the softmax layer and $s$ denotes the output label. The total sum of reward is computed after analyzing all the selected ROIs, as defined below,

$$R = \sum_{t=1}^{T} \gamma^{t-1} r_t \quad (6)$$

where $\gamma^{t-1}$ is the weight factor for reward $r_t$ at time t. For a finite horizon problem such as classification, we set $\gamma = 1$. Here the learning task is to optimize the parameters $\theta$ that maximize the overall performance $L(\theta)$, associated with reward $r_t$. A straightforward approach for handling this maximization task is by using the gradient ascent. The update rule follows the standard backpropagation and is defined as,

$$\theta_{n+1} = \theta_n + \alpha \nabla_\theta L(\theta_n) \quad (7)$$

where $n$ is the iteration index and $\alpha$ is the learning rate. In order to maximize $L(\theta)$, we employ the REINFORCE rule [31] from the class of policy gradients to adjust the model parameters. In this episodic scenario, on average, the model computes the gradients for actions that lead to higher rewards and consequently, the log probability of actions with low rewards will be decreased. The policy gradient, as described above, can be mathematically expressed as follows,

$$\nabla L_\theta = \sum_{t=1}^{T} \nabla_\theta \log \pi((l_{t+1}, Y_t) \mid (i_t, I^{\downarrow 16}); \theta) R_t \quad (8)$$

or

$$\nabla L_\theta = \sum_{n=1}^{N} \sum_{t=1}^{T} \nabla_\theta \log \pi((l_{t+1}^n, Y_t^n) \mid (i_t^n, I^{\downarrow 16}); \theta) R_t^n \quad (9)$$

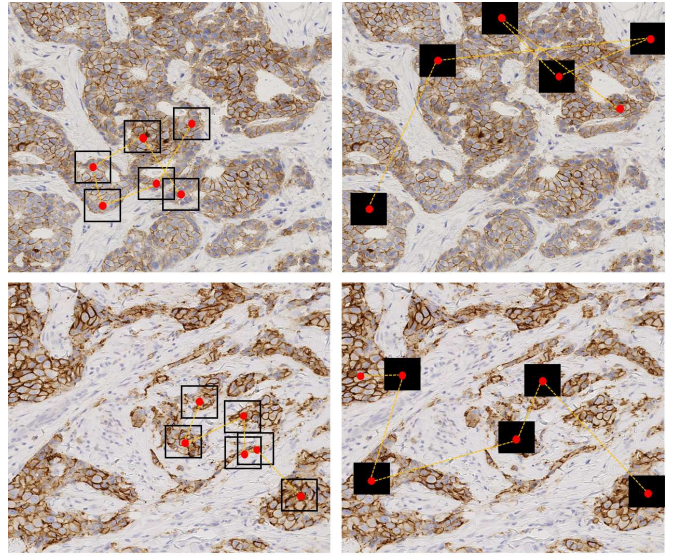

Fig. 4: Two sample images (*Top* and *Bottom*) representing the effect of inhibition of return (IoR), showing the selected ROIs without the IoR penalization (*Left*) and with the IoR penalization (*Right*). As can be seen, the selected ROIs after the IoR penalization are relatively distinct from each other. Filled rectangular regions (black) show the suppressed texture.

One limitation of the above formulation is that the model convergence can be challenging if intra-class variance in the training dataset is relatively high. To generalize the policy gradient algorithm, we include a baseline function $b_t = \mathbb{E}_\pi[R_t]$ for comparing the action values to the cumulative reward [32],

$$\nabla L_\theta = \sum_{n=1}^{N} \sum_{t=1}^{T} \nabla_\theta \log \pi((l_{t+1}^n, Y_t^n) \mid (i_t^n, I^{\downarrow 16}); \theta)(R_t^n - b_t) \quad (10)$$

*D. Inhibition of Return*

An important factor in adequate selection of diagnostically relevant regions is to inhibit the model from visiting the previously attended regions. We observe that for some of the selected locations during the sequential process, the sampled ROIs are not spatially distinct. Fig 4 (1st column) shows two such examples where the selected ROIs lie relatively close to each other, resulting in overlap with previously attended regions without any significant performance gain. This argument also applies to images where the diaminobenzidine (DAB) stain expression is relatively sparse as shown in Fig. 4

($2^{nd}$ row). A straightforward strategy to address this issue would be to suppress the texture information [8] of previously attended locations that would encourage the model to rapidly explore spatially distinct locations. This simple Inhibition of return (IoR) strategy leads to the model giving higher priority to regions that it has not previously considered for learning the discriminative features. The IoR strategy is computationally efficient, widely studied in cognitive psychology [33] and in sequential learning [8][18].

Further, we introduced an additional constraint $L_{IoR}$ in the loss function. At the end of the iteration sequence $t = (1,...,T)$, $L_{IoR}$ computes the overlap between the coordinates of selected ROIs, penalizing selection of image patches relatively close to each other. The scope of IoR penalization vanishes (its value becomes 0) if selected locations are spatially distinct from each other. The $L_{IoR}$ term is defined as below,

$$L_{IoR} = \frac{1}{C_2^T} \sum_{t=1}^{T} \sum_{j=t+1}^{T} rect[l_t(x,y)] \cap rect[l_j(x,y)] \quad (11)$$

where $rect[l_t]$ and $rect[l_j]$ represent rectangular coordinates sampled from the input image $I$ and $C_2^T$ is the number of combinations of different ROIs, in turn helping in limiting the intersection values between 0 and 1. The above loss function updates the policy parameters to correctly predict the HER2 score by penalizing the model for locating spatially overlapping regions.

*E. Task-Specific Regularization*

The clinical impact of large erroneous predictions of HER2 score is highly significant and should be avoided as much as possible. Inaccurate prediction for patients of score 0/1+ as 3+ will lead to giving treatment with toxic anti-HER2 drugs to patients who do not need it, while predicting cases with score 3+ as 0/1+ will lead to the patient not given the appropriate treatment needed. To avoid such scenarios, we added a task-specific regularization term.

$$L_{sc} = |\operatorname*{argmax}_{s}(Y_T') - g| \quad (12)$$

The final loss function combines the parameterized loss with task-specific regularization and IoR as given in (13),

$$L = L_\theta + \lambda (L_{sc} + L_{IoR}) \quad (13)$$

where $\lambda$ controls the sensitivity (scope of penalty) for both $L_{sc}$ and $L_{IoR}$.

IV. EXPERIMENTS AND RESULTS

*A. The Dataset*

This study is conducted on a publicly available dataset from the HER2 scoring contest [9]. The contest dataset consists of WSIs from 172 histology slides of 86 invasive breast carcinomas cases scanned using Hamamatsu NanoZoomer C9600 at the highest resolution (40 ×), two slides per case (one IHC stained with HER2 and another with the standard H&E). On average, each scanned WSI contains more than $10^{10}$ pixels. The GT for the contest was marked by a minimum of two expert histopathologists. For each case in the training dataset, the GT consists of a HER2 score and a percentage of complete membrane staining (PCMS), both at the WSI level. The training dataset is made of 52 cases, 13 cases from each HER2 score (0-3+) and the test set consists of 28 cases. The remaining 6 cases were not included in the test/training dataset and only reserved for the on-site part of the competition, see [12] for more details.

*B. Experimental Setup*

The ROIs were cropped at 40 × and 20 × resolutions, each ROI of size 128 × 128 × 3 pixels. The size of the input image $I$ was 2048 × 2048 × 3 (471.1 × 471.1 $\mu m^2$) sampled at 40 ×. In total, we extracted 58,500 tiles (with 50% overlap) from the 52 training WSIs after tissue segmentation and a simple DAB intensity based thresholding. The number of neurons in hidden layers of RNN was set to 256 and 128, respectively. The CNN transforms the given image into the feature representation of size 1 × 128. ReLU activation function was used after each residual block. To overcome the overfitting problem, we performed data augmentation by random rotations (0°, 90°, 180°, 270°), horizontal and vertical flipping, and the transpose of all the images in the training dataset. The regularization parameter $\lambda$ controls the sensitivity of the task-specific regularization and IoR penalization, as in (13). Through empirical observations, we found that the best performance was achieved with a value of 0.04 for $\lambda$, which we used for all the experiments. The initial learning rate was set to 0.001 with exponential reduction of 0.97 and the momentum was adjusted at 0.9. The batch size was selected as 10. The location of the first ROI was randomly selected. The number of ROIs per image was selected as 6, more details on this in Section IV.C.2. The learning parameters were initialized as Gaussian random numbers with 0 mean and $10^{-2}$ standard deviation and biases were set to 0.

*C. Comparative Analysis*

In this section, we discuss a variety of experiments to demonstrate the efficacy and evaluate the performance of the proposed method. For the following experiments, we performed 4-fold cross validation across 52 cases. We split the 52 cases into 4 subsets, with nearly equal representation of all four HER2 scores, and used 3 subsets for training and the remaining one subset for validation. The GT for the test dataset is not publicly available and, in this section, we have reported the performance of different variants of the proposed model on the validation dataset. We report the results for the test dataset in Section IV.D. Generally, a large part of WSI contains background (glass) regions with no tissue components. For tissue segmentation, we perform local entropy filtering on a lower resolution (2.5 ×) version of the WSI.

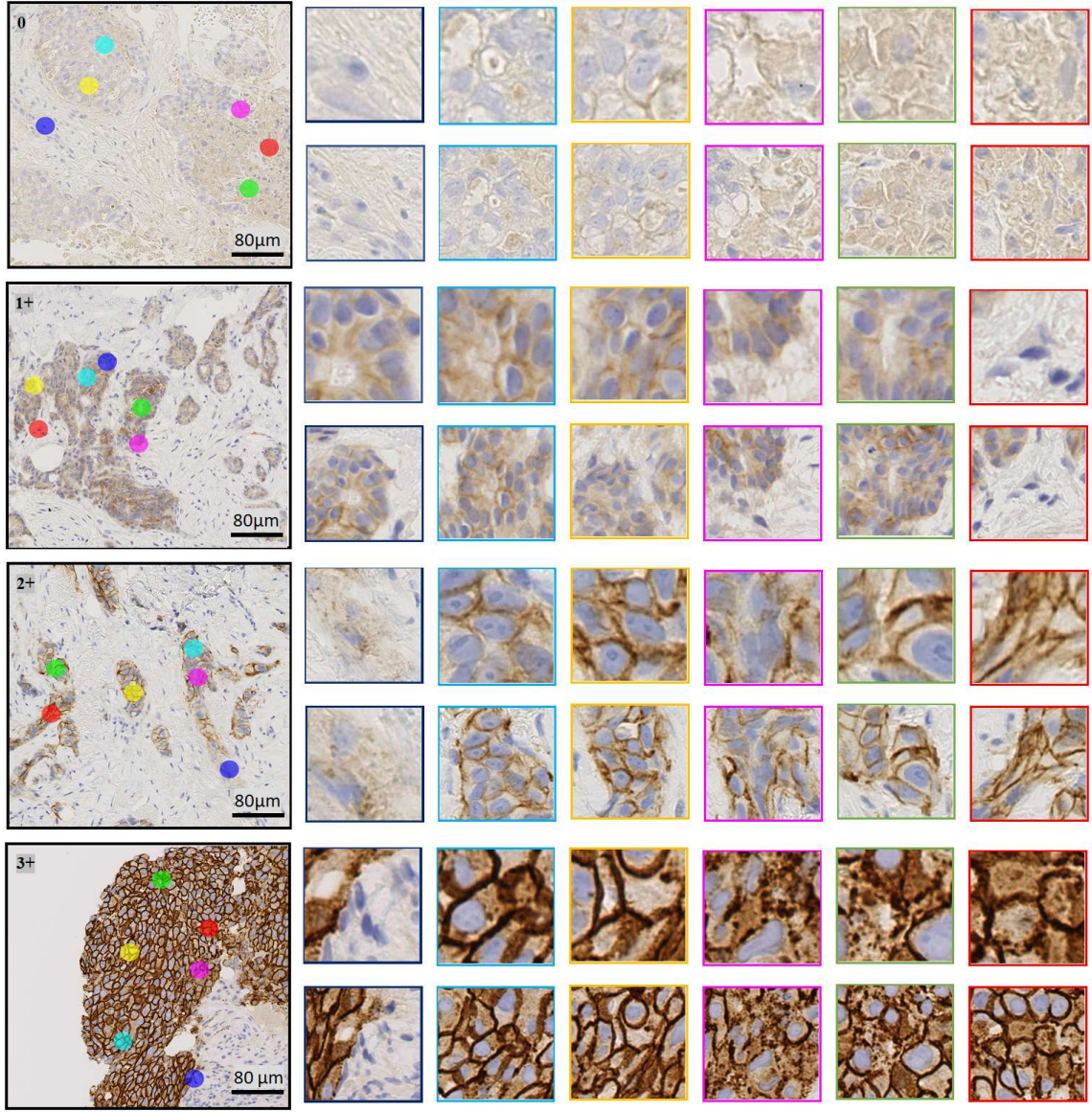

Fig. 5: Example of four image tiles with selected regions-of-interest (ROIs) predicted by our method, for each HER2 score (0-3+), respectively. The first column shows the input images and colored disks shows the predicted locations. The remaining columns show the selected regions at 40× and 20× around the selected locations $l_t$, $t = 1, 2, \ldots, 6$. The first selected region is shown with blue bounding box and the last selected region is shown with red bounding boxes.

*1) Comparative Results*: In this experiment, we evaluate the significance of different sub-components of the proposed model, including $L_\theta$, $L_{sc}$, and $L_{IoR}$. Another important aspect is to evaluate the significance of the parameterized policy for selecting relevant ROIs and how it affects the performance if we select ROIs randomly instead of following a certain policy. For random selection of ROIs, we perform two main experiments: a) select ROIs randomly from the entire $I$ and b) select ROIs randomly from only DAB regions of $I$. In the first approach of random selection, we predict the HER2 score of given ROIs ($i_t$) and select the next location randomly without consulting the context and current state of the model. Fig. S-1 in Supplementary Material-I shows the qualitative results for randomly selected regions and locations selected by the parameterized policy. Random selection correctly predicts HER2 score for images where most of the area is covered by discriminative tissue regions. However, it is susceptible to selecting regions that contain mostly background and sparse DAB regions. For the second approach, we perform stain deconvolution [34] on $I$ by estimating the stain matrix using [35]. DAB regions contain low luminance and therefore for binarizing the DAB channel using $\tau(I_{DAB})$, we empirically chose a relatively high threshold value of 0.8. The $\tau(I_{DAB})$ is then followed by morphological operations to exclude the noisy (small) components of the DAB channel. Table I shows the results for both experiments. Evidently, the second method is a relatively direct way of selecting $i_t$ and therefore yields higher performance as compared to the first approach for random selection. However, it offers some limitations as compared to the proposed model: the first major limitation is the absence of $L_{IoR}$, that enables the model to explore spatially distinct locations. The $L_{IoR}$ restrains the proposed model from

TABLE I
COMPARATIVE RESULTS

| Method | 0 | 1+ | 2+ | 3+ | $Acc_{comb}$ |
|---|---|---|---|---|---|
| RMVA [31] | 0.702 | 0.446 | 0.275 | 0.275 | 0.355 |
| random ROIs, $L_\theta, L_{sc}$ | 0.868 | 0.671 | 0.632 | 0.803 | 0.743 |
| random DAB ROIs $L_\theta, L_{sc}$ | 0.822 | 0.615 | 0.677 | **0.874** | 0.764 |
| Proposed – without CM | 0.982 | 0.452 | 0.568 | 0.721 | 0.652 |
| Proposed – $L_\theta$ | **0.982** | 0.538 | 0.703 | 0.782 | 0.733 |
| Proposed – $L_\theta, L_{sc}$ | 0.963 | 0.532 | **0.721** | 0.825 | 0.753 |
| Proposed – $L_\theta, L_{sc}, L_{IoR}$ | 0.919 | **0.772** | 0.661 | 0.837 | **0.794** |

$Acc_{comb}$ denotes combined accuracy and CM (in 3$^{rd}$ row) denotes the context module. In DAB ROIs, the locations were randomly selected from the diaminobenzidine (DAB) regions.

overemphasizing on particular locations by penalizing the learnable parameters and encourages the model to learn discriminative features from different tissue regions. One way of handling the absence of $L_{IoR}$ is to introduce a set of hard-constraints for selecting spatially distinct $l_t(x, y)$. However, defining a set of generalized hard-constrained is a non-trivial task and it may influence the model performance on images where we have sparse DAB representation (3$^{rd}$ row of Fig. 5). Secondly, it has no longer access to the overall $I^{\downarrow 16}$ and the context module. Consequently, this variant of our proposed model could be considered a departure from the routine clinical practice, where a pathologist glances through a coarse representation of input image at a higher level and then selects ROIs at lower levels before concluding the outcome. Therefore, it is imperative to follow a parameterized policy that incorporates the context and offers a temporal connection (via LSTM) between the selected locations.

Further, we investigate the implications of the context module, task-specific regularization ($L_{sc}$) and IoR ($L_{IoR}$). The context module allows the model to analyze a coarse representation of the overall environment and use that to predict $l_{t+1}(x, y)$. Existing attention based models [23][36] including RMVA (recurrent model for visual attention) have no mechanism to prevent models from revisiting the previously attended regions. Besides, RMVA also incorporates a strong location prior to the model, which is irrelevant in histology image analysis mainly due to the random orientation and morphological appearance of underlying tissue. Table I gives patch-based classification results of the proposed model with different settings. Overall the results are in favor of the proposed method ($L_\theta, L_{sc}, L_{IoR}$). Fig. 5 shows a sample of representative patches with selected ROIs. The most challenging images were from HER2 class 1+ and 2+, where resemblance in morphological appearance was present due to tumor heterogeneity, as shown in 2$^{nd}$ and 3$^{rd}$ rows of Fig. 5. In Fig. 5, the image patch for score 2+ contains relatively smaller extent of DAB expression and most of the tissue region belongs to 0/1+. In that case, the model starts with a relatively less informative region and sequentially learn to focus on the informative regions of the image to predict the correct outcome.

*2) Number of ROIs:* The aim of this experiment is to investigate the effect of number of ROIs for a given image $I$. We evaluate the performance of the proposed model by using

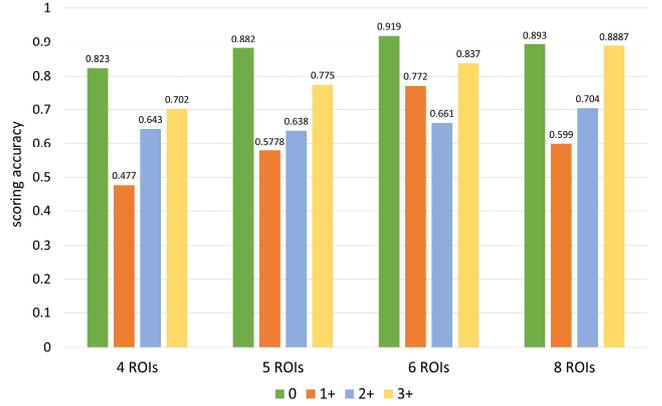

Fig. 6: Comparative results for different numbers of ROIs.

4, 5, 6 and 8 number of ROIs. For this contest dataset, we observed the best performance with 6 ROIs, as shown in Fig. 6. One of the main reasons for relatively low performance with 8 ROIs is the images containing tissue boundary regions, where most of the image region is covered by background glass. Therefore, in those cases, selecting more locations may confound the model in predicting the correct outcome. Another interesting aspect is the reduction of inference time and gain in the overall performance (as discussed in Section IV.D), in a conventional patch-based setting, for a given image $I$ of size $2048 \times 2048 \times 3$ at $40 \times$. The model needs to process all 64 ROIs (each of size $256 \times 256 \times 3$). In contrast, the proposed model can predict the HER2 score after consulting a handful of ROI patches accompanied with down-sampled version of $I$. However, although computing time may not be the most decisive aspect in clinical practice, it may be an important factor in high-volume diagnostic settings and for high-throughput IHC screening.

TABLE II
SIGNIFICANCE OF CONTEXT

| Method | 0 | 1+ | 2+ | 3+ | $Acc_{comb}$ |
|---|---|---|---|---|---|
| Proposed $40 \times, 20 \times$ | **0.919** | **0.772** | 0.661 | **0.837** | **0.794** |
| Proposed $40 \times, 10 \times$ | 0.881 | 0.613 | **0.676** | 0.807 | 0.742 |
| Proposed $20 \times, 10 \times$ | 0.802 | 0.592 | 0.608 | 0.711 | 0.65 |

$Acc_{comb}$ – combined accuracy

*3) Selection of multiple resolution regions:* This experiment compares the performance of the proposed model for selecting a suitable combination of magnification levels. We perform this experiment with three sets of magnification levels including $40 \times, 20 \times$, and $10 \times$. We observed that for IHC HER2 scoring, ROIs selected from $40 \times$ and $20 \times$ yield the best performance. The results for mean scoring accuracy for all 4 classes are shown in Table II. ROIs selected at higher resolutions offer more detailed information regarding HER2 expression at cell levels. On the other hand, ROIs selected at $10 \times$ or lower magnification offer more context information, but they are also more likely to contain background or irrelevant tissue regions, including non-invasive hematoxylin stained regions.

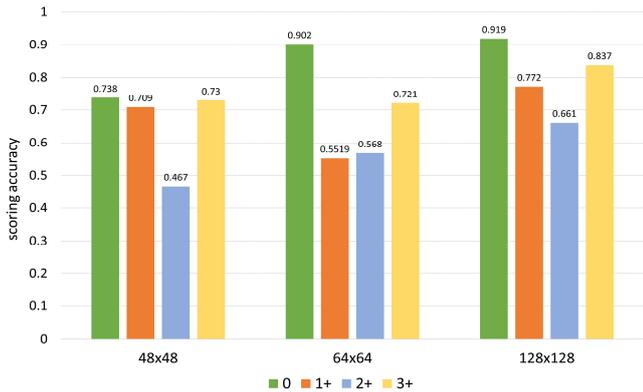

Fig. 7: Comparative results for different patch sizes of ROIs.

*4) Size of ROI:* The main objective of this experiment was to evaluate the performance of the proposed model on different sizes of ROIs. After selecting the location $l_t(x, y)$, it is worthwhile to quantify the extent of context required for predicting the correct outcome. Recent studies in computational pathology have also emphasized the significance of visual context [37]. We perform this experiment on three different patch sizes, including $48 \times 48, 64 \times 64,$ and $128 \times 128$, with results in Fig. 7. We noticed the best performance with ROIs of size $128 \times 128$. As expected, the performance of the proposed model increases with increase in the context. However, in HER2 scoring, it is important to limit the ROI size to prevent the inclusion of irrelevant tissue regions.

### D. Contest Leaderboards:

This subsection covers the description for scaling the patch level results to WSIs and performance of the proposed algorithm on the contest tasks.

*1) Contest Tasks:* The performance of the proposed algorithm on the WSI level is evaluated by using 3 different criteria, as suggested in the contest guidelines: a) agreement points, b) weighted confidence and c) combined points. In *agreement points*, a penalty method was introduced that assigned points between 0 and 15 to each case based on the clinical significance of the difference in predicted and actual scores. To resolve tie situation in the first criterion, bonus points were also awarded based on a correct prediction of PCMS. A *weighted confidence* was devised to estimate the credibility of WSI results predicted by the algorithm. This measure may also help in stratifying cases that need further input from pathologists. And finally, for each case, a *combined point* was calculated by taking the product of the other two assessment criteria. Further details regarding the evaluation criteria is explained in [9].

*2) PCMS Estimation:* In routine practice, a pathologist visually estimates the PCMS on the WSI level, indicating the strength of invasive carcinoma cells stained to HER2 protein. For our experiments, we split the WSI into manageable image tiles $I$, depending on the computational resources. The model then predicts a HER2 score for each image tile and aggregates the results on the WSI level by simply choosing the most dominant class as the HER2 score $s = argmax \, (s_0; s_{1+}; s_{2+}; s_{3+})$ where $s_0; s_{1+}; s_{2+}; s_{3+}$ represent the number of image tiles predicted as $0, 1+, 2+, and \, 3+,$ respectively. The PCMS for invasive breast cases was estimated by the WSI output maps. For each score map, we simply compute the ratio between the area covered by each predicted score over the total area of tissue region within the WSI.

TABLE III
COMPARATIVE RESULTS

| Teams | Pts | Pts+B | Cf | W.Pts |
|---|---|---|---|---|
| **The proposed method** | **405** | **419** | **24.1** | **359.1** |
| VISILAB-I (GoogleNet [38]) | 382.5 | 404.5 | **23.55** | 348 |
| FSUJena [15] | 370 | 392 | 23 | 345 |
| HUANGCH (AdaBoost) | 377.5 | 391.5 | 22.62 | 345.7 |
| MTB NLP (AlexNet [39]) | 390 | 405.5 | 22.94 | 335.7 |
| VISILAB-II (contour analysis) | 377.5 | 391 | 21.88 | 322 |
| Team Indus (LeNet [40]) | **402.5** | **425** | 18.45 | 321.4 |
| UC-CCSE [17] | 390 | 395 | 21.07 | 316 |
| MUCS-III [11] | 390 | 411 | 20.43 | 300.8 |
| HERcules (SVM) | 360 | 380 | 20.57 | 295.6 |
| MUCS-II (GoogleNet [38]) | 385 | 413 | 19.51 | 290.1 |

Description of the notations used above, Pts: points, Pts+B: points with bonus, Cf: weighted confidence, and W.Pts: combined leaderboard for weighted points and points with bonus.

*3) Combined Leaderboards:* One of the most challenging aspects in analyzing histology images is to limit the automated analysis to diagnostically relevant tissue parts.—A fully supervised method processes the given image regardless of noisy or irrelevant contents. In contrast, recurrent models can appropriately tackle such scenarios by only processing the relevant ROIs. Fig. S-2 (Supplementary Section-II) shows some of the visual results on the image patches from the validation dataset. Table III reports the WSI level performance of the proposed method on all 3 evaluation criteria. It also contains the results of top-10 performing algorithms in the contest. The proposed method ranked as $1^{st}$ amongst all 18 submissions in the contest, including the combined points criterion and point based scoring. The contesting algorithms were based on a wide range of the state-of-the-art deep convolutional networks including GoogleNet [38], AlexNet [39] and LeNet [40]. On 28 WSIs from the test dataset, the proposed algorithm correctly classified 26 WSIs. The most difficult cases were from borderline class 2+ and 3+. Both the false predictions belonged to those classes. It is worth noting that the scores of 2+ and 3+ are the most difficult to call for expert histopathologists as well. On the other hand, the proposed model correctly predicted the scores of 12 out of 14 WSIs with scores 2+ or 3+.

### E. Glyoxalase-1 protein (Glo1) Scoring:

In this experiment, we evaluate the performance of the proposed method on IHC stained gastroenteropancreatic neuroendocrine tumors (GEPs). The over-amplification in glyoxalase-1 protein (Glo1) is associated with resistance in multidrug tumor chemotherapy [41]. In routine practice, an expert pathologist visually examines IHC stained GEP slides and reports a score between 1+ and 3+, which represents weak, moderate and intense Glo1 immunostaining, respectively. This experiment is conducted on 82 WSIs from 39 patients with 25 midguts and 14 pancreatic cases, with a total number of 22, 33 and 27 WSIs scored by an expert pathologist as 1+, 2+ and 3+ respectively. Further details regarding the dataset and GEP tumors can be found in [42].

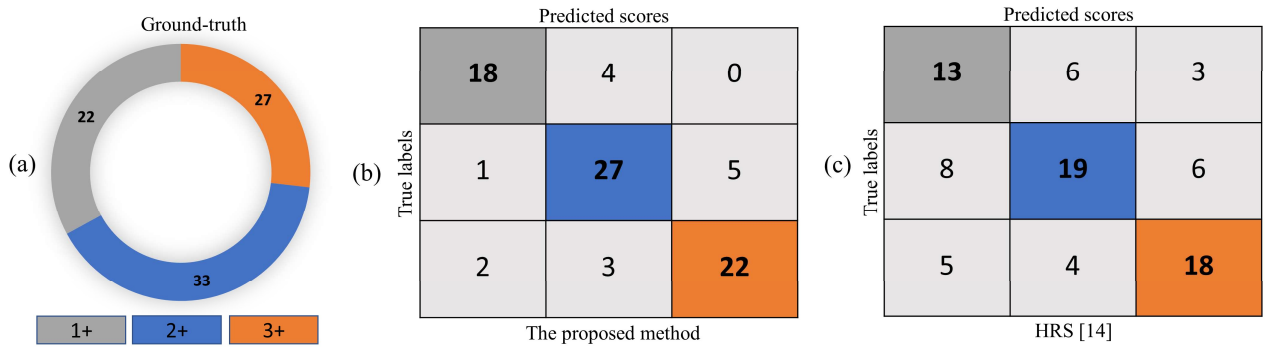

Fig. 8: (a) Description of the number of Glo1 scoring WSIs (b) confusion matrix for the proposed method and (c) confusion matrix for the hormone receptors scoring (HRS) method [14].

The main objective of this experiment is to test the efficacy of the proposed IHC scoring method on another IHC stain by quantifying the concordance between the pathologist GT score and Glo1 score predicted by the proposed method. We first split the dataset into two folds and perform cross-validation by selecting half of the dataset for training and the remaining half for testing. We then repeat the experiment by swapping the training and test datasets. For each fold, the training dataset was further spilt into training and validation subsets by selecting 35 WSIs for training and remaining 6 (≈15% of the training data) for validation. Tissue regions from WSIs were segmented by performing local entropy filtering on a lower resolution (2.5×) version of the WSI. To avoid the class imbalance problem, we extracted all the image tiles (containing tissue) from 1+ WSIs and randomly sampled equal number of image tiles from classes 2+ and 3+. In total, we extracted 86,700 tiles each of size $2048 \times 2048 \times 3$ at 40×. We then retrained the proposed model by retaining the same data augmentation methods and hyperparameters (including learning rate, lambda, number of ROIs, etc.,) as explained in Section IV.B. Optimal weights were selected based on the validation set. The final Glo1 score on the WSI level was estimated by aggregating the scores of image tiles, as explained in Section IV.D.2.

Fig. 8 shows two confusion matrices (CF) containing WSI results of Glo1 scoring. The CF in Fig. 8(b) shows the results for the proposed method and the results for automated hormone receptors scoring (HRS) method [14], which was initially used in [42], are shown in Fig. 8(c). The overall agreement between the GT and the proposed method is 81.7% (Glo1 scores for 67 WSIs were correctly predicted), whereas the agreement is 60.9% in case of HRS. For IHC scoring, HRS relies heavily on pixel intensities for quantifying the chromatin and protein content. Therefore, heterogeneous morphological characteristics of neuroendocrine cells within tumor regions pose the risk of confusing the algorithm in extracting the desired features for IHC scoring. Fig. 9 shows some of the qualitative results on image tiles with different Glo1 scores. Overall, it is encouraging that the proposed method outperforms the HRS with a noticeable margin. It is worth mentioning that the performance of the proposed method may improve by appropriately tuning the hyperparameters. Our intention here is to demonstrate to some extent the generalizability of the hyperparameters selected from HER2 scoring. The IHC scoring of HER2 and Glo1 cases have some fundamental similarities: a) in both cases the GT was provided on the WSI level and therefore, it is imperative that the underlying model learns a stochastic policy and identifies some of the diagnostically relevant regions in predicting the final outcome, and b) similar to HER2 scoring, an erroneous scoring of 1+ as 3+ or vice versa may have far reaching effects for a patient. It is somewhat necessary to have a mechanism that penalizes the learnable parameters for overcalling those classes. We also observed that in some tissue regions, cells are densely packed and pose difficulties for the model in selecting ROIs from the invasive tumor regions.

## V. DISCUSSION

With the adoption of digital slide scanners in routine pathology labs, large-scale WSIs or virtual slides ($10^{10}$ pixels) have emerged as a reliable alternative to conventional glass slides [43]. Typically, for effective training of deep learning models, it is incumbent to train computational models with large-scale precisely annotated regions. The ineludible fact is that sourcing precise high-resolution annotations of scanned slides is a laborious task and not considered as a part of the routine clinical practice. Hence, attaining tissue-level annotations for a significantly large dataset is one of the factors that may delay the acceptability of automated methods in clinical practice [44].

This study proposes a novel recurrent model for IHC scoring of HER2 slides of breast cancer. In some respects, the model mimics the pathological behavior by learning a parameterized policy to select diagnostically relevant regions. Experimental results conducted on a challenging contest dataset demonstrate the efficacy of the proposed model by outperforming the state-of-the-art methods, as shown in Table III. It is worth noting that the proposed model only identifies a small number of regions required for predicting the correct outcome. Depending on the requirements, the model can be extended to assign different scalar rewards to each selected

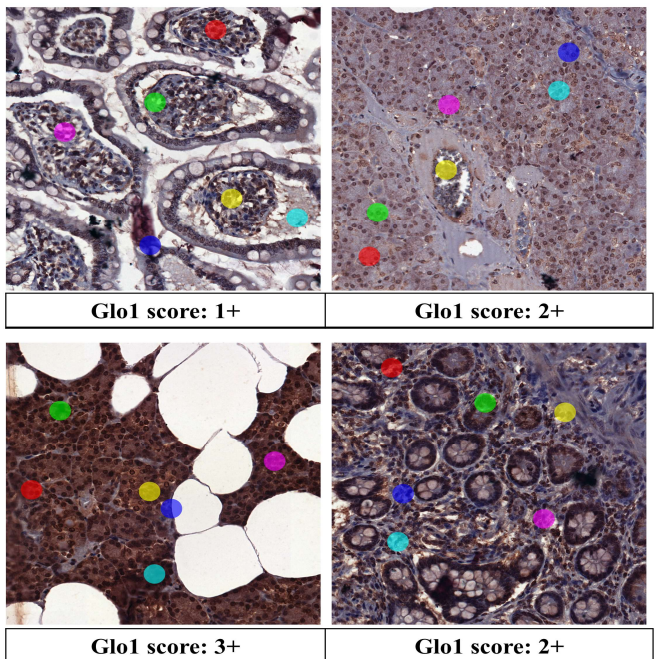

Fig. 9: Sample of four IHC stained Glo1 images with selected regions-of-interest (ROIs) predicted by our method. Colored disks show the predicted locations and the sequence is as follows: blue, cyan, yellow, magenta, green, and red.

ROI. Different rewards may also be interpreted as the significance of each selected ROIs. Nevertheless, a challenge remains in that some times the model selects regions that may not appear to be diagnostically relevant (e.g., non-invasive tumor regions) due to tumor heterogeneity. An example is shown in Fig. 5, the last location of score 1+.

In computational pathology, diagnostic efficiency may be based on two main factors: 1) selecting potential ROIs from a given WSI, and 2) extracting discriminative features from within the selected ROIs. Some studies on eye movement have shown that with the passage of time trainee pathologists eventually gain experience in the selection of diagnostically relevant ROIs and learn discriminative features. Another recent study [45] shows that the overlap ratio between ROIs selected by different pathologists is associated with higher diagnostic accuracy. An interesting extension of our work would be to find correlation between the diagnostically relevant ROIs selected by machines and experts. This may help in evaluating the diagnostic reliability of computer-assisted diagnostic systems. The proposed model also offers several advantages over the conventional fully supervised approaches. First, the model is capable of handling unwanted background regions that are often quite common in histology images due to several non-standardization factors in slide preparation. Second, the proposed model is capable of scaling up to WSIs or significantly large tiles of a WSI as the number of trainable parameters are directly linked with the size of ROIs instead of image tiles. However, in order to perform end-to-end training of the proposed model, we need the uncompressed representation of all required tiles of a WSI to be hosted into the memory and it is worth mentioning that on average, a 24-bit representation of an uncompressed WSI at 40 × requires approximately 56 GB [46] of memory, making it intractable to load an entire WSI into the GPU memory. Another extension of our work to achieve end-to-end learning would be to devise a multi-stage (or scale) attention mechanism. In the first stage, model analyzes the given WSI at a lower magnification (1.25 × or lower) to identify the potential ROIs and in the second stage the attention model analyzes selected ROIs at higher resolution (40×, 20 ×) to learn the discriminative features. However, such a model would require a prudent selection of reward function(s) and tuning of hyper-parameters to ensure the inclusion of diagnostically relevant regions in the first stage. In addition, the model would require a large number of WSIs to effectively optimize the weights for the first stage of the attention model.

The scope of the proposed model is not only limited to HER2 antibody quantification. The scoring of other prognostic and predictive markers like estrogen receptor (ER), progesterone receptor (PR), and proliferation of Ki-67 on the WSI level may also be possible with a similar approach. In the United States alone this year (2018), it is expected that there will be 266,120 new invasive BCa cases[1]. Whereas visual examination of histological specimens is generally influenced by subjectivity measures for learning discriminative features. the proposed model may assist experts in reducing subjectivity and serve as a semi-automated tool to find diagnostically relevant regions that may require the pathologist attention.

## VI. CONCLUSIONS

In this study, we presented a deep reinforcement learning approach for automated scoring of IHC stained HER2 slides of breast cancer. Unlike fully supervised models that process all the regions of a given input image, the proposed model treats IHC scoring as a sequential selection task and effectively localizes diagnostically relevant regions by deciding *where to see*. The proposed model carries the potential to solve other histology image analysis problems where it is difficult to get precise pixel-level annotations. A similar approach may also eventually assist the pathologist in automated localization and classification of potential ROIs in both H&E and IHC stained histology images. The evaluation was conducted on a publicly available HER2 scoring contest dataset. Detailed comparative evaluation is in the favor of the proposed model. This study could potentially commence more interpretable incorporation of sequential learning and policy gradients in the domain of computational pathology.

[1] https://www.breastcancer.org/symptoms/understand_bc/statistics

SUPPLEMENTARY MATERIAL – I

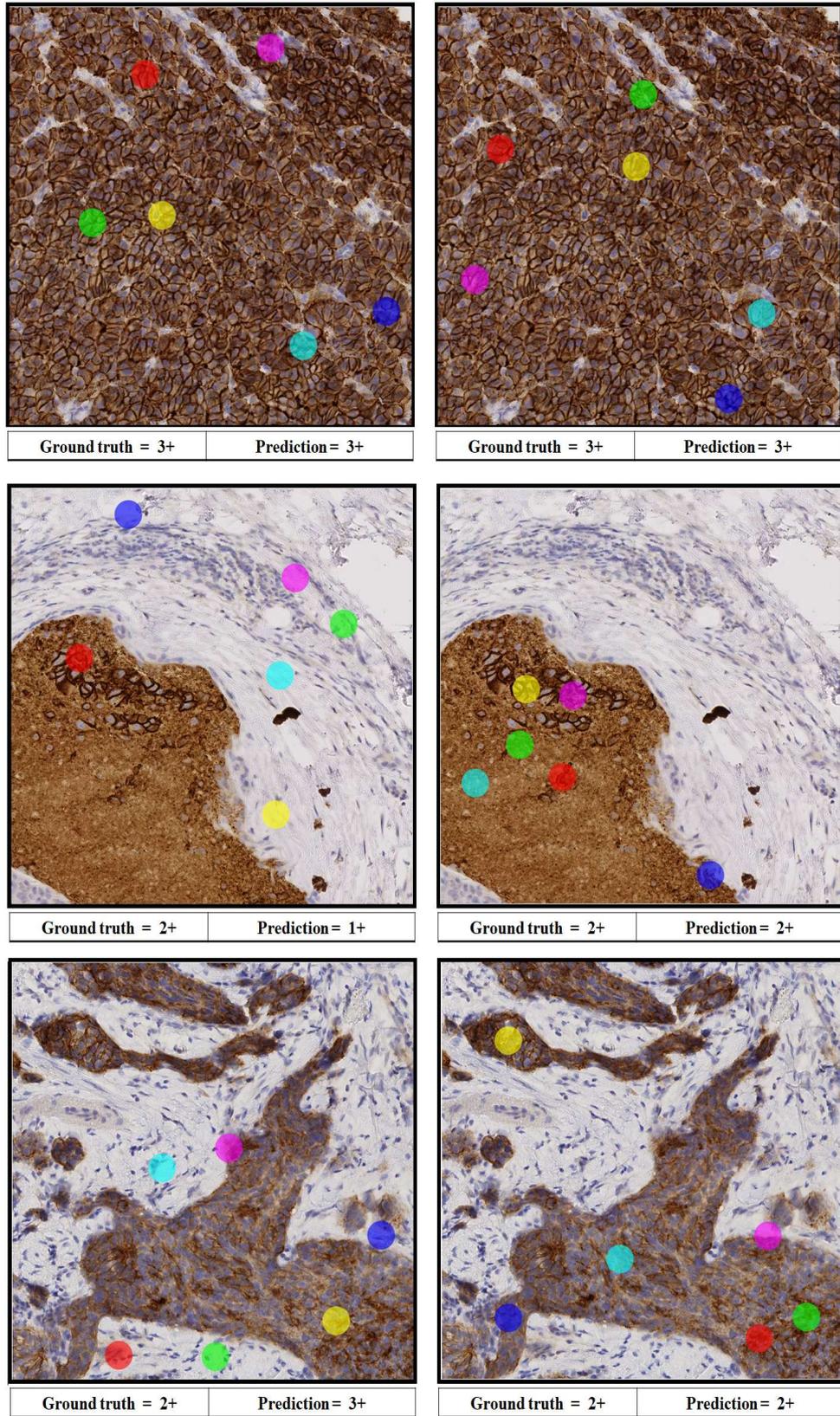

Fig. S-1: *(left)* randomly sampled locations *(right)* location sampled by following the parameterized policy. Colored circles show the predicted locations and the sequence is as follows blue, cyan, yellow, magenta, green, and red.

Supplementary Material - II

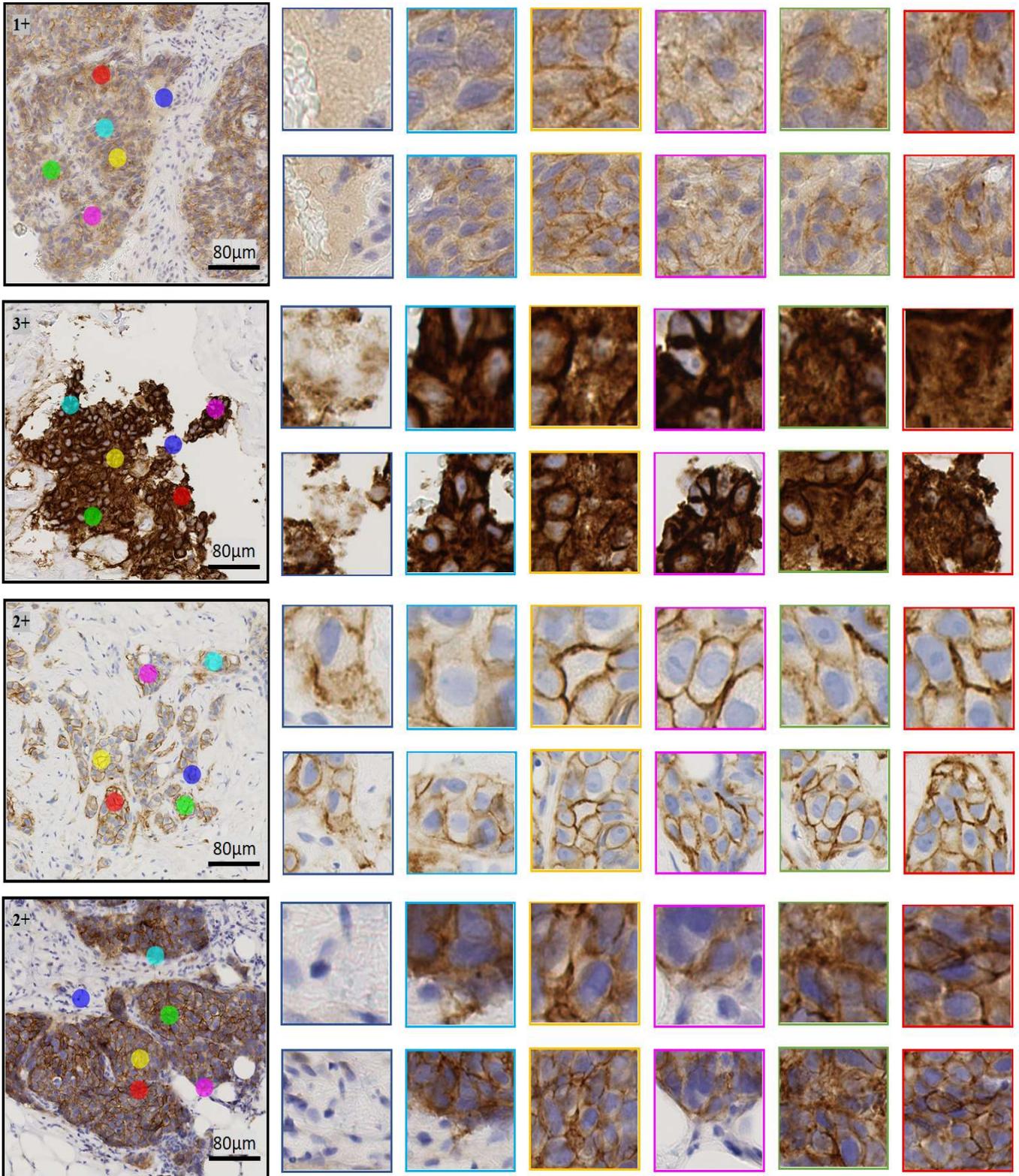

Fig. S-2: Examples of four images with selected regions-of-interest (ROIs) predicted by the algorithm. Correctly predicted HER2 scores are 1+, 3+, 2+, and 2+. The first column shows input images and colored circles shows the predicted locations. The first location is shown with blue bounding boxes and similarly the last location is shown with red bounding boxes.